\documentclass{amica-paper-biorxiv}

\usepackage{booktabs}
\usepackage{longtable}
\usepackage{multirow}
\usepackage{url}
\usepackage[ruled,vlined]{algorithm2e}
\usepackage{xurl} 

\newcommand{\code}[1]{\texttt{\nolinkurl{#1}}}

\DontPrintSemicolon
\SetKwInput{KwInput}{Input}
\SetKwInput{KwOutput}{Output}
\SetAlCapNameFnt{\small\sffamily\bfseries}
\SetAlCapFnt{\small\sffamily}
\crefname{algocf}{Algorithm}{Algorithms}
\Crefname{algocf}{Algorithm}{Algorithms}

\leadauthor{Huberty}

\begin{document}

\title{AMICA-Python: Adaptive Mixture Independent Component Analysis with Anderson Acceleration}
\shorttitle{AMICA-Python}

\author[1]{Scott Huberty}
\author[2]{Christian O'Reilly}
\affil[1]{University of Southern California, Los Angeles, CA, USA}
\affil[2]{University of South Carolina, Columbia, SC, USA}

\maketitle

\begin{abstract}
Adaptive Mixture Independent Component Analysis (AMICA) is widely used in EEG research
and has long been associated with strong empirical performance for blind source
separation. Despite its impact, practical use has historically depended on a single
Fortran implementation, accessed via the EEGLAB toolbox for MATLAB, limiting its
accessibility for analytical pipelines not designed within the MATLAB ecosystem. Here we present AMICA-Python,
a Python implementation of the AMICA algorithm, with a scikit-learn-conformant
API designed for integration with existing scientific Python pipelines. The
implementation follows the reference algorithm closely while adopting modern software
engineering practices and an interface familiar to Python users. Additionally, we introduce
an optional Anderson acceleration scheme that can dramatically reduce the time to convergence
for this relatively slow algorithm \cite{anderson1965iterative, anderson2011Walker}.
To evaluate numerical agreement and practical performance, we benchmarked AMICA-Python
against the reference Fortran implementation on 14 open EEG recordings. After averaging 3 runs of each implementation on all 14 recordings, AMICA-Python closely matched the reference, with a median final normalized log-likelihoods of 11.572
for both the Fortran and Python implementations, and a negligible median relative absolute difference of only $1.07\times10^{-8}$ when normalized by the absolute Fortran value. Runtime was also competitive. Relative to the reference implementation, AMICA-Python was 17.7\% faster, while the Anderson-accelerated variant was 34.1\% faster. AMICA-Python reproduces the reference implementation to high numerical precision with competitive runtime, while making AMICA available
through a more accessible and extensible Python interface.
\end{abstract}


\begin{corrauthor}
seh33\at uw.edu
\end{corrauthor}

\section*{Introduction}
Independent component analysis (ICA) is a widely used tool for addressing the blind source
separation problem. Its applications are many; ranging from neuroscience, 
genetics, biology, chemistry, astronomy, seismology, stock analysis, text analysis,
and more
\cite{
      varoquaux_2010_groupica,%
      delorme_independent_2012,%
      pearlson_2015_pica,%
      rutledge_independent_2013,%
      lee_ica_genetics_2003,%
      scholz_ica-fingerprinting_2004,%
      cardoso_smica,%
      xiong_separation_2024,honkela_wordica_2010%
      }.
Linear ICA algorithms (which are currently more widely used than the non-linear variants \cite{hyvarinen_nonlinear_2023}),
assume that the observed multivariate data can be modeled as a linear mixture
of latent components. In other words, the observed data and the aforementioned latent
components are linked via a linear transform:

$$
X = A S
$$

\noindent where the observed data $X \in \mathbb{R}^{N \times T}$ contain $N$ features
(i.e. "channels" in the literature related to electroencephalography, or EEG) and $T$ observations (i.e. "samples" or "time points" in the EEG literature).
The $N \times N$ mixing matrix $A$ is unknown, and the matrix $S$ of latent components
(henceforth referred to as 'sources', not to be confused in the EEG literature with cortical/sub-cortical sources of the scalp EEG) contains rows that are maximally statistically
independent.

In EEG, in particular, ICA is routinely used to separate neural
activity from ocular, muscular, and other artifact sources \cite{huberty_pylossless_2026, vigario-ica}. Among available ICA
methods, Adaptive Mixture Independent Component Analysis (AMICA) \cite{palmer_amica_2011}
is particularly favored within the EEG research community, primarily because of its
empirical performance on several benchmarks \cite{delorme_independent_2012, leutheuser_comparison_2013}.
For example, in a systematic comparison of 22 ICA and blind source separation algorithms,
AMICA produced both the largest reduction in pairwise mutual information between recovered
components and the largest percentage of ``near-dipolar'' components, a property associated
with physiologically plausible cortical brain activity \cite{delorme_independent_2012}.

\begin{figure}[t]
\centering
\includegraphics[width=\columnwidth]{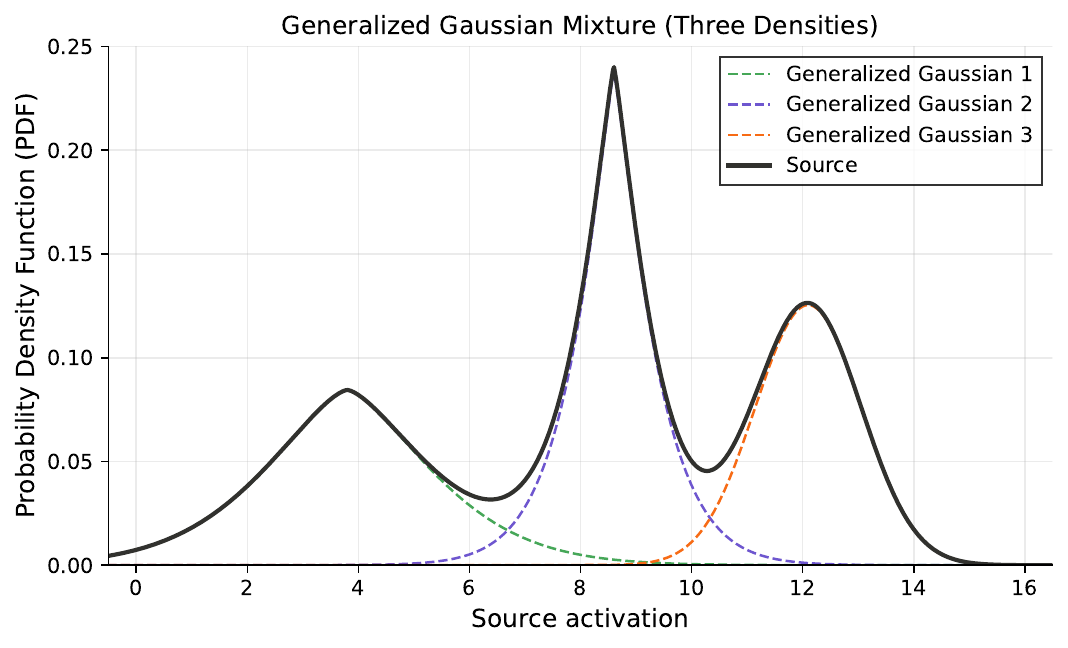}
\caption{Illustration of mixture of a generalized Gaussians.}
\label{fig:gmm-plot}
\end{figure}

AMICA is a maximum-likelihood ICA method in the Infomax lineage
\cite{bell_information-maximization_1995, lee_independent_1999}. 
Extended Infomax can identify sources with sub-Gaussian (lighter tail) and super-Gaussian (heavier tail) distributions, via
nonlinear functions that are derived from pre-specified source probability density 
functions (or in other words, fixed score functions; see Ille (2024) for a description
\cite{ille_orthogonal_2024}). In contrast, AMICA treats the
source distributions themselves as parameters to be learned, by modeling each source as a
mixture of generalized-Gaussian densities (\cref{fig:gmm-plot}), where the generalized-Gaussian shape parameter $\rho$ controls
tail heaviness and peakedness:

\begin{equation} \label{eq:gg}
p(x;\mu,\beta,\rho)
=
\frac{\rho\beta}{2 \Gamma(1/\rho)}
\exp\left[
-\left(\beta|x-\mu|\right)^\rho
\right].
\end{equation}

Here, $\rho = 2$ corresponds to the Gaussian case, $\rho < 2$ gives
super-Gaussian, more heavy-tailed densities  (e.g., $\rho = 1$ is the Laplace density), and $\rho > 2$ gives sub-Gaussian,
more light-tailed densities. The Fortran and Python
implementations by default bound $1 \leq \rho \leq 2$, so the component family interpolates between Laplace-like and Gaussian densities.

In this adaptive approach,
the estimates of the source distributions are updated alongside the unmixing matrix during
optimization. As a result, AMICA can match a broader class of source distributions,
rather than committing in advance to a predefined set of distributions
(e.g. sub- and super-Gaussian). While AMICA was not the first ICA algorithm
to propose such an approach \cite{te-won_lee_ica_2000}, this is the central conceptual
difference between AMICA and ICA methods such as Infomax and extended Infomax.

The flexibility of AMICA comes at a cost. Generally speaking, ICA methods with
gradient-based optimization (including Infomax) are already known to converge slowly,
at least compared to fixed-point algorithms such as FastICA \cite{Montoya_caveats_2017, hyvarinen_independent_2000}.
A number of accelerations have been proposed to improve convergence speed
of these algorithms \cite{ablin_faster_2017, ille_orthogonal_2024},
and AMICA itself adopts a Newton step along the lines of Amari (1997) to speed up
convergence \cite{amari_stability_1997, Palmer_newton_2008}. Still, the
combined cost of jointly estimating the source densities and the unmixing matrix at each
iteration make AMICA much more computationally expensive than algorithms like Extended
Infomax, both in terms of convergence behavior and per-iteration computation time
\cite{Frank2023, ablin_faster_2017}.

These considerations have shaped AMICA's software history. The reference
implementation is a Fortran program that was originally published in 2011 \cite{amica}.
As Lulkin and colleagues point out, a pure MATLAB version of the AMICA algorithm was also created by
the algorithm author \cite{Lulkin-AMICA-Julia}, however this version has no official
publication or distribution, and Lulkin and colleagues further demonstrated that it can
be about 8x slower than the Fortran implementation \cite{Lulkin-AMICA-Julia}. For this
reason, the Fortran program remains the canonical reference implementation. Further, it
is accessed almost exclusively through the EEGLAB MATLAB toolbox \cite{delorme_eeglab_2004}.
While this continues to serve the MATLAB-based EEG community well, it places AMICA outside the
ecosystem in which much modern ICA work is now done.

In contrast, ICA algorithms such as FastICA and Extended-Infomax are available as
pure-Python libraries that integrate with scikit-learn, MNE-Python, and
related tools \cite{scikit-learn, ablin_faster_2017}. Similarly,
spectral matching ICA, which is preferred by astrophysics researchers, has been ported
to Python for use in adjacent communities \cite{ablin_spectral_2021}. For most of its
history, AMICA has lacked a comparable and competitive implementation, which complicates
its inclusion in analytical pipelines, and raises the barrier to studying or modifying
the algorithm itself. Advances in both hardware and software have made carrying out previously burdensome
scientific computations feasible in more user-friendly, interpreted programming
languages such as Python. For example, Lulkin (2023) present a re-implementation of AMICA
in the Julia programming language \cite{Lulkin-AMICA-Julia}. However, the Python ecosystem
would benefit from a robust and actively maintained version of the AMICA algorithm, as Python is an extremely popular
programming language for both machine learning and EEG research
\cite{mne_python_software, scikit-learn}. While pure Python is known to be relatively
slow for scientific programming, much of this limitation is mitigated by its most widely used packages
for numerical processing (e.g., Numpy, Scipy, Pandas, Polars), which utilize compiled
extensions to low-level languages such as C, Fortran, or Rust to carry out
computationally heavy operations \cite{harris2020array, 2020SciPy-NMeth,
mckinney-proc-scipy-2010, polars2026}. This is one reason Python remains so popular
for scientific computing and machine learning analyses.

The present work describes AMICA-Python, a Python implementation of the AMICA algorithm
that (i) exposes a scikit-learn-compatible estimator interface that can be used within
scikit-learn pipelines, (ii) is validated numerically against the reference Fortran
implementation on matched inputs, and (iii) is profile-optimized to remain tractable on
realistic datasets (e.g. long EEG recordings that may be 1GB or more on disk). We
show that we can produce an even more performant implementation of AMICA via careful
algorithmic implementation and the use of modern numerical libraries. However, we also
optionally allow AMICA-Python users to further improve convergence time by utilizing
Anderson Acceleration of AMICA, as we will describe below. The sections that follow
summarize the AMICA model and the Damped Anderson Acceleration Method for fixed-point
iteration algorithms, describe the design of AMICA-Python, and report correctness and
performance comparisons against the Fortran reference.

\subsection*{The AMICA model}

Comprehensive descriptions and notation of the AMICA algorithm have been presented in
multiple publications. First, Palmer and colleagues described the method in a series of
papers \cite{Palmer_super-gaussian_2006, Palmer_newton_2008, palmer_amica_2011}, followed
by expositions by Hsu (2018) and Lulkin (2023) \cite{hsu_amica_2018, Lulkin-AMICA-Julia}.
The following sections give a brief summary of the AMICA approach in an
attempt to provide intuition and facilitate readers' understanding. Our discription differs from previous publications in two ways: First, the notation here is
aligned with the Fortran and Python implementations. We present several
equations in log space, because it matches more closely the numerical implementation in
Fortran and AMICA-Python, which operate in log-space for the purpose of numerical stability. Thus, our notation
is aligned with the implementation, even though previous descriptions of AMICA typically 
presented the generalized-Gaussian mixture model without the inclusion of logarithmic transforms. Second, we describe the steps for performing a single decomposition with AMICA (i.e., a single-model AMICA), whereas the Fortran program optionally allowed for simultaneously fitting multiple decompositions at a time. Doing so requires additional operations across a "model" axis, however, we will not describe these steps in the notation in order to simplify the description and facilitate the readers understanding of the core algorithm.


\begin{table}[t]
\centering
\caption{Key symbols used in the AMICA description. Indices $i$,
$j$, and $t$ run over components, mixture terms, and samples, respectively.
The sphering matrix $Q$ is returned as \texttt{S} by the implementation;
see text.}
\label{tab:notation}
\begin{tabular}{ll}
\toprule
Symbol & Meaning \\
\midrule
$X^{\mathrm{raw}}$ & observed data in sensor or feature space \\
$Q$ & sphering matrix \\
$X$ & mean-centered, sphered data used by AMICA \\
$A$ & mixing matrix in sphered data space \\
$W$ & unmixing matrix in sphered data space, $W=A^{-1}$ \\
$\mathbf{c}$ & fitted AMICA bias in sphered data space \\
$\mathbf{Y}$ & recovered source estimates \\
$u_{ijt}$ & scaled, centered mixture argument \\
$\beta_{ij}$ & scale parameter \\
\bottomrule
\end{tabular}
\end{table}

\subsection*{Mixture Densities}

Let $X^{\mathrm{raw}} \in \mathbb{R}^{N \times T}$ denote the observed data matrix
with $N$ channels and $T$ samples. AMICA first mean-centers and whitens the data (i.e., applies sphering). In
the full-rank setting considered here, applying the sphering matrix
$Q \in \mathbb{R}^{N \times N}$ gives
\begin{equation}
X \;=\; Q\cdot \bigl(X^{\mathrm{raw}} - \bar{\mathbf{x}}\mathbf{1}^{\!\top}\bigr)
\;\in\; \mathbb{R}^{N \times T},
\label{eq:sphered-data}
\end{equation}
where $\bar{\mathbf{x}} \in \mathbb{R}^{N}$ is the channel-wise mean and
$\mathbf{1} \in \mathbb{R}^{T}$ is a vector of ones. Optionally, AMICA-Python can retain only
$K \leq N$ sphered components, but the notation below assumes the full-rank case.

AMICA then estimates a nonsingular mixing matrix
$A \in \mathbb{R}^{N \times N}$ in sphered data space, together with
$W=A^{-1}$ and a fitted bias vector $\mathbf{c} \in \mathbb{R}^{N}$. The recovered
sources are
\begin{equation}
\mathbf{y}_t \;=\; W\mathbf{x}_t - W\mathbf{c},
\qquad t = 1, \dots, T,
\label{eq:amica-unmixing}
\end{equation}
or, equivalently, $\mathbf{Y}=W(X-\mathbf{c}\mathbf{1}^{\!\top})$. Here,
$\mathbf{c}$ is a center in sphered data coordinates, so $W\mathbf{c}$ is the
corresponding source-space offset.
The vector $\mathbf{y}_t \in \mathbb{R}^{N}$ denotes the current estimates of all
$N$ sources at sample $t$.

As in other ICA methods, the decomposition is only defined up to a permutation and
scaling of the recovered sources. AMICA fixes part of this ambiguity by normalizing
columns of the sphered-space mixing matrix $A$ during optimization and absorbing the
associated scale changes into the source-density parameters \cite{palmer_amica_2011, hsu_amica_2018}.

As previously mentioned, the key idea in AMICA is that the source densities are learned
rather than fixed in advance. In AMICA, each recovered source is
modeled by a mixture of one-dimensional generalized-Gaussian densities~\cite{Palmer_super-gaussian_2006, Palmer_newton_2008, palmer_amica_2011, hsu_amica_2018}.

In plain terms, the density of each recovered source is modeled as a weighted
sum of one-dimensional mixture components, each parameterized by its own
weighted mixture of generalized Gaussians with their own location, inverse scale, and shape parameters, as defined in eq. (\ref{eq:gg}).



The per-source mixture-density parameters are gathered into matrices
$\boldsymbol{\alpha}, \boldsymbol{\mu}, \boldsymbol{\beta},
\boldsymbol{\rho} \in \mathbb{R}^{N \times M}$, where $M$ is the number of
density mixtures per source. Here $\alpha_{ij}$ is a mixture weight,
$\mu_{ij}$ is a location, $\beta_{ij}$ is an inverse scale parameter, and
$\rho_{ij}$ is a shape parameter. With these in hand, we can define the per-mixture log-density
matrix $\mathbf{Z} \in \mathbb{R}^{N \times M \times T}$. Let:

\begin{equation}
q(u;\rho)
=
\frac{\rho}{2\Gamma(1/\rho)}
\exp\left(-|u|^\rho\right)
\label{eq:unit-generalized-gaussian}
\end{equation}

denote the unit-scale generalized-Gaussian density evaluated at the standardized source value $u$.
Then the scaled component density in \cref{eq:gg} can be written as
$\beta_{ij}q(\beta_{ij}(y_{it}-\mu_{ij});\rho_{ij})$:

\begin{equation}
Z_{ijt}
\;=\;
\log \alpha_{ij}
+ \log \beta_{ij}
+ \log q\!\bigl(\beta_{ij}(y_{it} - \mu_{ij});\rho_{ij}\bigr),
\label{eq:amica-mixture-logterm}
\end{equation}

We further derive $u_{ijt}$, the source value after centering by mixture component
$(i,j)$ and multiplying by that component's inverse scale as
\begin{equation}
u_{ijt}
\;=\;
\beta_{ij}\bigl(y_{it} - \mu_{ij}\bigr).
\label{eq:amica-standardized-source}
\end{equation}

This standardized value is the argument fed to the component density $q$.

In the AMICA-Python source code, the function \code{compute\_source\_densities} populates $\mathbf{Z}$ in a single
vectorized call. The mixture weights of each source satisfy
\begin{equation}
\sum_{j=1}^{M} \alpha_{ij} = 1.
\label{eq:mixture-weights}
\end{equation}

\subsection*{Source densities and mixture responsibilities}

Once $\mathbf{Z}$ has been populated, AMICA uses it in two operations
along the mixture axis (axis $j$). First, the log-density of each recovered source at
each sample is obtained by collapsing the mixture dimension with a $logsumexp$ (LSE) operation. Let
$\log \mathbf{P} \in \mathbb{R}^{N \times T}$ denote the resulting matrix of source
log-densities after summing over mixture densities:

\begin{equation}
\log \mathbf{P} \;=\; \mathrm{LSE}_j\bigl(\mathbf{Z}\bigr).
\label{eq:amica-density}
\end{equation}

\noindent where the index $j$ to the operator $\mathrm{LSE}$ indicates that the LSE operation is performed along the dimension of the index $j$. Second, normalizing $\mathbf{Z}$ along the same mixture axis with softmax
yields the responsibility tensor
$\mathbf{R} \in \mathbb{R}^{N \times M \times T}$:
\begin{equation}
\mathbf{R} \;=\; \mathrm{softmax}_j\bigl(Z\bigr),
\qquad
\sum_{j=1}^{M} R_{ijt} = 1\quad \forall\, i,t.
\label{eq:amica-estep}
\end{equation}

Each entry $R_{ijt}$ is the posterior probability that mixture component $j$
generated sample $t$ of source $i$. The two operations consume the same per-mixture
log terms but serve different purposes: $\mathrm{LSE}_{j}$ produces the
per-source log-density used in the likelihood, whereas $\mathrm{softmax}_{j}$
produces the responsibilities used to reweight the source-density parameter updates.

\subsection*{Likelihood and parameter updates}

The model parameters updated on each iteration are

\begin{equation}
\begin{aligned}
\Theta = \{&
W, \mathbf{c}, \boldsymbol{\alpha}, \boldsymbol{\mu},
\boldsymbol{\beta}, \boldsymbol{\rho}\}.
\end{aligned}
\end{equation}

The log-likelihood of one sample $t$ is the sum
of the sphering Jacobian, the unmixing Jacobian, and the per-source
log-densities from \cref{eq:amica-density}. The final term measures how well the
recovered sources fit their current density models:
\begin{equation}
\ell_t
= \log|\det Q|
  + \log|\det W|
  + \sum_{i=1}^{N} \log p_i(y_{it}),
\label{eq:amica-sample-ll}
\end{equation}

and the full log-likelihood is the sum over samples,
\begin{equation}
\mathcal{L}(\Theta) \;=\; \sum_{t=1}^{T} \ell_t.
\label{eq:amica-ll}
\end{equation}

Note that the Fortran implementation and AMICA-Python report the normalized log-likelihood,
\begin{equation}
\bar{\mathcal{L}}(\Theta)
\;=\;
\frac{1}{NT}\sum_{t=1}^{T} \ell_t,
\label{eq:amica-normalized-ll}
\end{equation}
which is the quantity used for convergence monitoring and the benchmark
comparisons below.

\paragraph{Convergence criteria.}

By default, both the Fortran implementation and AMICA-Python monitor
convergence using two criteria: the normalized log-likelihood and a weight-gradient norm.
Let $r$ index optimization iterations and define

\begin{equation}
\Delta\bar{\mathcal{L}}^{(r)}
\;=\;
\bar{\mathcal{L}}(\Theta^{(r)})
-
\bar{\mathcal{L}}(\Theta^{(r-1)}).
\label{eq:amica-ll-increment}
\end{equation}

AMICA also computes the root-mean-square norm of the current weight-update
matrix. In the single-model, full-rank case described here, if
$\Delta A^{(r)} \in \mathbb{R}^{N \times N}$ denotes the update direction
formed for the mixing matrix at iteration $r$, this quantity is
\begin{equation}
\eta^{(r)}
\;=\;
\frac{\|\Delta A^{(r)}\|_{\mathrm{F}}}{N},
\label{eq:amica-update-norm}
\end{equation}


\noindent where \(\|\cdot\|_{\mathrm{F}}\) denotes the Frobenius norm. With thresholds $\varepsilon_{\mathcal{L}}$ and $\varepsilon_g$, AMICA terminates when either
the normalized log-likelihood improves by less than
$\varepsilon_{\mathcal{L}}$ for the configured number of consecutive
iterations, or when the mixing-matrix update norm falls to or below
$\varepsilon_g$:

\begin{equation}
\Delta\bar{\mathcal{L}}^{(r)} < \varepsilon_{\mathcal{L}}
\quad\text{or}\quad
\eta^{(r)} \leq \varepsilon_g.
\label{eq:amica-convergence}
\end{equation}

\paragraph{Density parameter updates.}

The responsibility matrix $\mathbf{R}$ from \cref{eq:amica-estep} drives
updates for the mixture weights, locations, and scales.

For each source, mixture component, and sample, define the density penalty

\begin{equation}
f_{ij}(u_{ijt})
\;=\;
-\log q(u_{ijt};\rho_{ij})
\label{eq:amica-density-penalty}
\end{equation}

\noindent as the negative log-density assigned by component $(i,j)$ to the
standardized source value $u_{ijt}$. With this convention, the derivative of the density penalty used in the
source-density and unmixing updates is given by

\begin{equation}
f'_{ij}(u_{ijt})
\;=\;
\rho_{ij}\,\operatorname{sign}(u_{ijt})\,|u_{ijt}|^{\rho_{ij}-1}.
\label{eq:amica-density-penalty-derivative}
\end{equation}

The updates for mixture weights, locations, and scales (given $\rho_{ij} \leq 2$) are
\begin{align}
\alpha_{ij}^{+}
&= \frac{1}{T}\sum_{t=1}^{T} R_{ijt},
\label{eq:amica-alpha}\\
\mu_{ij}^{+}
&= \mu_{ij}
 + \frac{\sum_{t} R_{ijt}\,f'_{ij}(u_{ijt})}
        {\beta_{ij}\sum_{t} R_{ijt}\,f'_{ij}(u_{ijt}) / u_{ijt}},
\label{eq:amica-mu}\\
\beta_{ij}^{+}
&= \beta_{ij}
\left(
\frac{\sum_{t} R_{ijt}}
     {\sum_{t} R_{ijt}\,f'_{ij}(u_{ijt})\,u_{ijt}}
\right)^{1/2}.
\label{eq:amica-beta}
\end{align}

Conceptually, samples with larger responsibility for mixture component
$(i,j)$ contribute more strongly to that component's next location and
inverse scale, which is how the source-density model adapts to the data on every
iteration. 

Finally, the shape parameters $\rho_{ij}$ have no closed-form and
are updated by a scaled gradient step \cite{Palmer_super-gaussian_2006}:
\begin{equation}
\begin{aligned}
\rho_{ij}^{+}
&=
\rho_{ij}
+ \eta_{\rho}
\Biggl[
  1
  - \frac{\rho_{ij}}{\psi\!\left(1+\tfrac{1}{\rho_{ij}}\right)}
\\
&\qquad\qquad
  \times
  \frac{\sum_{t} R_{ijt}\,|u_{ijt}|^{\rho_{ij}}\log|u_{ijt}|^{\rho_{ij}}}
       {\sum_{t} R_{ijt}}
\Biggr].
\end{aligned}
\label{eq:amica-rho}
\end{equation}

where $\psi$ denotes the digamma function (i.e., the derivative of the log of the gamma function) and $\eta_{\rho}$ is the
shape-parameter learning rate.

\paragraph{Unmixing-matrix update.}

For each recovered source $i$ and sample $t$, AMICA combines the
mixture-component density penalty derivatives into one derivative
$g_i(y_{it})$. Each mixture component $j$ contributes according to its
responsibility $R_{ijt}$:
\begin{equation}
g_i(y_{it})
\;=\; \sum_{j=1}^{M} R_{ijt}\,\beta_{ij}\,f'_{ij}(u_{ijt}),
\label{eq:amica-score}
\end{equation}
so that $\mathbf{G} \in \mathbb{R}^{N \times T}$ collects $g_i(y_{it})$
entry by entry; in AMICA-Python, this quantity is computed by the function
\code{compute\_scaled\_scores}. With $\mathbf{G}$ and the recovered-source
matrix $\mathbf{Y} \in \mathbb{R}^{N \times T}$, the natural-gradient update
of $W$ takes the matrix form
\begin{equation}
\Phi
\;=\; \tfrac{1}{T}\,\mathbf{G}\,\mathbf{Y}^{\!\top}
\;\in\; \mathbb{R}^{N \times N},
\qquad
\Delta W \;\propto\; (I - \Phi)\,W.
\label{eq:amica-natgrad}
\end{equation}

The matrix $\Phi$ summarizes the empirical relationship between the current
source-level density penalty derivatives and the recovered sources. The natural-gradient ascent
update is structurally identical to Infomax and extended Infomax
\cite{bell_information-maximization_1995, amari_stability_1997}, but
$\mathbf{G}$ is now data-adaptive because it is derived from the learned
source densities rather than from a fixed nonlinearity. In the Fortran
implementation and AMICA-Python, the corresponding step is applied to the
mixing matrix $A$, after which $W=A^{-1}$ is recomputed. Thus, each
iteration alternates between refining the source-density model and
refining the unmixing matrix so that the two estimates improve together.

\subsection*{Adaptive Newton Method}

Natural-gradient optimization is effective, but it is not especially fast. AMICA
augments the gradient update with a Newton method~\cite{Palmer_newton_2008} that exploits the curvature information of the gradient to guide
the unmixing matrix update. For the present paper, the main point is simply that
AMICA uses the currently estimated source statistics to construct better-scaled updates for $W$, which can improve
convergence once the source-density model is already a reasonable fit to the
data.

\subsection*{Algorithm summary}

The single-model AMICA procedure used in the present benchmark is summarized
in Algorithm~\ref{alg:single-model-amica}. At a high level, the algorithm alternates between improving
the source-density model and improving the unmixing estimate, so that each
informs the other.

\begin{algorithm}[t]
\caption{Single-model AMICA}
\label{alg:single-model-amica}
\KwInput{Observed data $X^{\mathrm{raw}}$, number of mixtures $M$,
maximum iterations $R_{\max}$, tolerances
$\varepsilon_{\mathcal{L}}$ and $\varepsilon_g$}
\KwOutput{Fitted parameters
$\Theta=\{W,\mathbf{c},\boldsymbol{\alpha},\boldsymbol{\mu},
\boldsymbol{\beta},\boldsymbol{\rho}\}$ and recovered sources $\mathbf{Y}$}
Mean-center and sphere the data:
$X \leftarrow Q(X^{\mathrm{raw}}-\bar{\mathbf{x}}\mathbf{1}^{\!\top})$\;
Initialize $A$, set $W\leftarrow A^{-1}$, and initialize
$\boldsymbol{\alpha},\boldsymbol{\mu},\boldsymbol{\beta},\boldsymbol{\rho}$\;
\For{$r=1,\ldots,R_{\max}$}{
Compute sources:
$\mathbf{Y}\leftarrow W(X-\mathbf{c}\mathbf{1}^{\!\top})$\;
Compute per-mixture log terms $\mathbf{Z}$ using \cref{eq:amica-mixture-logterm}\;
Compute source log-densities
$\log p_i(y_{it})\leftarrow\mathrm{LSE}_{j}(Z_{ijt})$\;
Compute responsibilities
$R_{ijt}\leftarrow\mathrm{softmax}_{j}(Z_{ijt})$\;
Compute $\bar{\mathcal{L}}^{(r)}$ and the weight-gradient norm $\eta^{(r)}$\;
\If{$\Delta\bar{\mathcal{L}}^{(r)}<\varepsilon_{\mathcal{L}}$
\textbf{or} $\eta^{(r)}\leq\varepsilon_g$}{
\textbf{break}\;
}
Update
$\boldsymbol{\alpha},\boldsymbol{\mu},\boldsymbol{\beta}$
using \cref{eq:amica-alpha,eq:amica-mu,eq:amica-beta}\;
Update $\boldsymbol{\rho}$ using \cref{eq:amica-rho}\;
Form source-level density penalty derivatives $\mathbf{G}$ using
\cref{eq:amica-score}\;
Update $A$ using the natural-gradient direction
\cref{eq:amica-natgrad} or Newton acceleration\;
Normalize columns of $A$ and absorb the scale into
$\boldsymbol{\mu}$ and $\boldsymbol{\beta}$\;
Refresh $W\leftarrow A^{-1}$\;
}
\end{algorithm}

\subsection*{Acceleration of the AMICA algorithm}

The AMICA iteration can be viewed as a fixed-point algorithm analogous to the Expectation-Maximization (EM). If
$\theta$ denotes the free AMICA parameters, one pass through the AMICA update
defines
\[
\theta_{k+1}=G(\theta_k),
\]
where $G$ includes the source-density updates, the mixing-matrix update, and the
normalization steps summarized above. Such algorithms are often stable but
slow, and this is one reason AMICA can be substantially more expensive than even Extended
Infomax: the method may require many iterations, and each iteration involves costly
matrix operations over samples, channels, sources, and mixture terms
\cite{delorme_independent_2012, Frank2023}.

Several methods have been proposed to accelerate the convergence of EM algorithms, both
in the context of ICA and for EM algorithms more generally. For example, Ablin and colleagues
significantly reduced the convergence time of Extended-Infomax by leveraging L-BFGS
optimization with backtracking line-search \cite{ablin_faster_2017}. 
General-purpose accelerators have also been proposed, which attempt to improve
convergence for any EM algorithm without deriving a problem-specific optimizer \cite{saad_acceleration_2025}.
For example, SQUAREM  is one such popular accelerator \cite{varadhan_simple_2008}. In the
context of AMICA, the problem with these approaches is that they typically require
multiple evaluations of $G$ per acceleration cycle. This is a poor trade-off when $G$ is
the expensive AMICA update.

However, in 2019, Varadhan and Henderson proposed an accelerator built on Anderson
optimization \cite{anderson1965iterative, anderson2011Walker} which is well suited to AMICA. Aptly named damped Anderson
acceleration with restarts and monotonicity control (DAAREM) \cite{henderson_damped_2019}.
Anderson acceleration forms an extrapolated next iterate from recent fixed-point
residuals $f_k=G(\theta_k)-\theta_k$. Rather than moving only to $G(\theta_k)$, it
solves a small least-squares problem over the last $m_k$ residual differences and uses
that solution to combine recent steps. The \textit{restarted} version used by DAAREM periodically
builds this history from order one up to a maximum order $m$, then discards the history
and starts the cycle again.

\begin{algorithm}[t]
\caption{Restarted Anderson acceleration}
\label{alg:restarted-anderson}
\KwInput{Initial parameter vector $\theta_0$, fixed-point map $G$, maximum order $m$}
\KwOutput{Accelerated fixed-point iterates}
Set $c_1\leftarrow 1$ and $\theta_1\leftarrow G(\theta_0)$\;
\For{$k=1,2,\ldots$ until convergence}{
Set $m_k\leftarrow \min(m,c_k)$ and $f_k\leftarrow G(\theta_k)-\theta_k$\;
Construct
$\mathcal{X}_k=[\Delta\theta_{k-m_k},\ldots,\Delta\theta_{k-1}]$
and
$\mathcal{F}_k=[\Delta f_{k-m_k},\ldots,\Delta f_{k-1}]$\;
Solve
$\gamma^{(k)}=\arg\min_{\gamma\in\mathbb{R}^{m_k}}
\|f_k-\mathcal{F}_k\gamma\|_2^2$\;
Propose
$\theta_{k+1}\leftarrow
\theta_k+f_k-(\mathcal{X}_k+\mathcal{F}_k)\gamma^{(k)}$\;
\If{$k \bmod m=0$}{
$c_{k+1}\leftarrow 1$\;
}
\Else{
$c_{k+1}\leftarrow c_k+1$\;
}
}
\end{algorithm}

DAAREM adds two safeguards to the restarted Anderson loop in
\cref{alg:restarted-anderson}. First, it damps the least-squares coefficients by solving
a ridge-regularized version of the Anderson problem; the amount of damping is adaptive,
starting close to the original fixed-point update and approaching the undamped Anderson
step as recent proposals are accepted. Second, it applies an
$\varepsilon$-monotonicity check against a merit function, usually the log-likelihood:
if the extrapolated candidate decreases the merit function by more than $\varepsilon$,
the algorithm rejects the extrapolation and falls back to the ordinary fixed-point
iterate. At the end of each restart cycle, DAAREM also checks whether the merit function
has decreased over the cycle and, if so, increases damping for subsequent proposals
\cite{henderson_damped_2019}.

A major advantage of DAAREM as compared to previously mentioned accelerators is that it
does not require additional evaluations of the $G$ per iteration. Anderson mixing has
already been shown to be useful for estimating the components of Gaussian mixture models,
and in machine learning more broadly \cite{sun2021damped, walker2011anderson}.
When applied to AMICA, we show that DAAREM further improves convergence behavior while
minimally impacting the decomposition performance.

\section{Methods}
\subsection*{AMICA-Python software}

AMICA-Python was designed as a native scientific Python implementation of the
single-model AMICA algorithm. At the user level, the package exposes a
scikit-learn-inherited estimator class, \texttt{AMICA}, whose API follows the familiar
\texttt{fit}, \texttt{transform}, and \texttt{fit\_transform} pattern
\cite{scikit-learn}, allowing the direct use of AMICA-Python in Scikit-Learn pipelines. After
fitting, the estimator exposes quantities expected by downstream arrays, including the
learned components, mixing matrix, whitening operator, and log-likelihood history.
AMICA-Python also includes helpers for interoperability with MNE-Python, allowing EEG
researchers to transfer the learned decompositions into MNE-Python for further analysis
and visualization \cite{mne_python_software}. In keeping with scikit-learn conventions,
input data are stored in \texttt{(n\_samples, n\_features)} layout rather than the
\texttt{channels} $\times$ \texttt{samples} notation used in the mathematical
description above and in the Fortran implementation.

The AMICA-Python user-interface is simplified as compared to its Fortran counterpart.
The Fortran program API exposes 86 tunable parameters, 4 of which are required to run
the program. Further, there are 36 parameters that contain boolean or enumerated arguments,
each of which trigger its own specific code path in the program. To simplify the user experience and reduce maintainer burden, AMICA-Python greatly reduces
the number of parameters in its estimator, and instead aims to provide sensible default
values for many of the hyper-parameters that the Fortran program exposes. From the users
end, they only need to pass the data to fit as a required argument to \texttt{AMICA}.

The reference Fortran program is highly optimized for memory and therefore
relies heavily on nested loops over samples, components, and mixture terms.
That design is understandable, because AMICA is relatively demanding in both
computation and memory. Several intermediate steps, including source-density
evaluation and responsibility estimation, require temporary arrays that scale
with \texttt{n\_samples}, \texttt{n\_components}, and
\texttt{n\_mixtures}, meaning that these intermediate arrays can be \texttt{n\_mixtures}
times larger than the input data. In a Python implementation, directly reproducing this structure would be inefficient, because repeatedly iterating over large numerical arrays
in pure Python carries substantial interpreter overhead.
AMICA-Python therefore favors vectorized array operations and delegates the numerically
intensive work to PyTorch. This design makes better use of PyTorch's compiled C
extensions. To keep the implementation tractable on realistic M/EEG recordings, AMICA-Python
also supports batch-wise processing along the sample axis, allowing large recordings to be processed without materializing all intermediate arrays in memory at once. For sufficiently large inputs, this batch-wise mode is enabled automatically.

As previously mentioned, the current implementation uses PyTorch as its primary numerical
backend, which provides a uniform API across CPU and GPU devices. This was primarily a
pragmatic design choice, as local benchmarks showed PyTorch to be more performant than
Scipy and Numpy. However, as we will discuss in the \textit{Future directions} section, 
we are considering broadening the support for other numerical backends to cater to a
wider user base.

The AMICA-Python package is validated against the reference Fortran outputs with unit
tests, and this testing suite is integrated to the software's continuous integration/development workflow, to ensure
that continued development does not come at the cost of mathematical drift with the
reference implementation. Further, AMICA-Python features extensive documentation of
its API and several examples and tutorials to assist users.
Finally, AMICA Python is published to PyPi and Conda-forge, the main package indexes for the Python community.

\subsection*{Benchmark Analysis}
To benchmark AMICA-Python against the reference Fortran implementation, we ran both 
implementations on an open access EEG dataset containing 14 recordings
\cite{delorme_independent_2012}, each containing 71 channels. Each recording was fit with
three mixture terms per source for up to 2000 iterations. Both implementations were run
using equivalent parameters and in the same environment, e.g. on a CPU with 16Gs of RAM,
and limited to a single thread, run on the same hardware. For fairness, the initial unmixing matrix weights and density
parameters (locations, inverse scales, weights) were saved from the Fortran run and used in the
Python runs by loading the same initial $W$, $\beta$, and $\mu$ values. For one of the test recordings
(\texttt{gv84}), visual inspection indicated that a long terminal segment of the recording
was dominated by noise. We therefore retained the first 665 epochs for both the Fortran
and AMICA-Python runs, applying this truncation before creating the data matrices passed to either implementation.

For each recording, the benchmark recorded wall-clock fit time, the
normalized log-likelihood trajectory, the final normalized log-likelihood, and the number of
completed iterations. The benchmark results were then used to generate the aggregate
plots included here. For reference, the benchmark results and scripts are available at
the links below, where the commit hashes in the second column reflect the exact version
of the source code at the time that this benchmark was generated.

\begin{table}[t]
\centering
\caption{Repository versions used for benchmark generation.}
\label{tab:benchmark-commits}
\begin{tabular}{ll}
\hline
Repository & Commit \\
\hline
\href{https://github.com/scott-huberty/amica-python}{amica-python} &
\texttt{d1350ee0} \\
\href{https://github.com/scott-huberty/amica-benchmark}{amica-benchmark} &
\texttt{11b1b48} \\
\href{https://github.com/scott-huberty/amica-python-paper}{amica-python-paper} &
\texttt{18137ce} \\
\hline
\end{tabular}
\end{table}

\subsection*{Replication of the Delorme source-dipolarity analysis}
As an additional validation against the original AMICA benchmark, we reproduced the
mutual-information-reduction and source-dipolarity analysis from Delorme et al.
\cite{delorme_independent_2012}. the MATLAB scripts, 14 EEG recordings, and published results (algorithm decompositions, mutual information summaries) were released alongside the paper, as well as 3rd party toolboxes that were used (e.g. DIPFIT, an EEGLAB tool for fitting equivalent dipole models to ICA components) residual-variance
estimates used to generate the published comparison. We created a working copy of the original
release, added AMICA-Python as an additional decomposition, and reran the original
MATLAB mutual-information script to compute AMICA-Python's mutual information reduction
in the same format as the published algorithms.

To estimate source dipolarity for AMICA-Python, we ran the original scripts shared by Delorm and colleagues on the AMICA-Python decomposition files. Recording \texttt{gv84} was excluded from this
analysis to match the set of recordings used by the original plotting script. For each algorithm, the mean mutual information reduction across the remaining 13
recordings was plotted. Source dipolarity was summarized as the percentage of
components with DIPFIT residual variance below 5\%. The final figure was generated
from these same reduced coordinates, with AMICA-Python added as an additional point.

\section{Results}

Across the 14 datasets, AMICA-Python closely matched the final
normalized log-likelihoods of the reference implementation. Across the 14 recordings, after averaging the three runs per recording, AMICA-Python closely matched the final normalized log-likelihoods of the reference implementation. The median final normalized log-likelihood was 11.57285 for Fortran and 11.57285 for AMICA-Python. The median relative absolute difference was $1.07\times10^{-8}$, with a mean of
$4.93\times10^{-7}$ and a range from $2.79\times10^{-9}$ to $3.51\times10^{-6}$.  Relative differences were computed as

\begin{equation}
\frac{
    \left|\mathrm{LL}_{\mathrm{Python}} -
    \mathrm{LL}_{\mathrm{Fortran}}\right|
}{
    \left|\mathrm{LL}_{\mathrm{Fortran}}\right|
}
\label{eq:relative-ll-difference}
\end{equation}

\begin{figure}[t]
\centering
\includegraphics[width=\columnwidth]{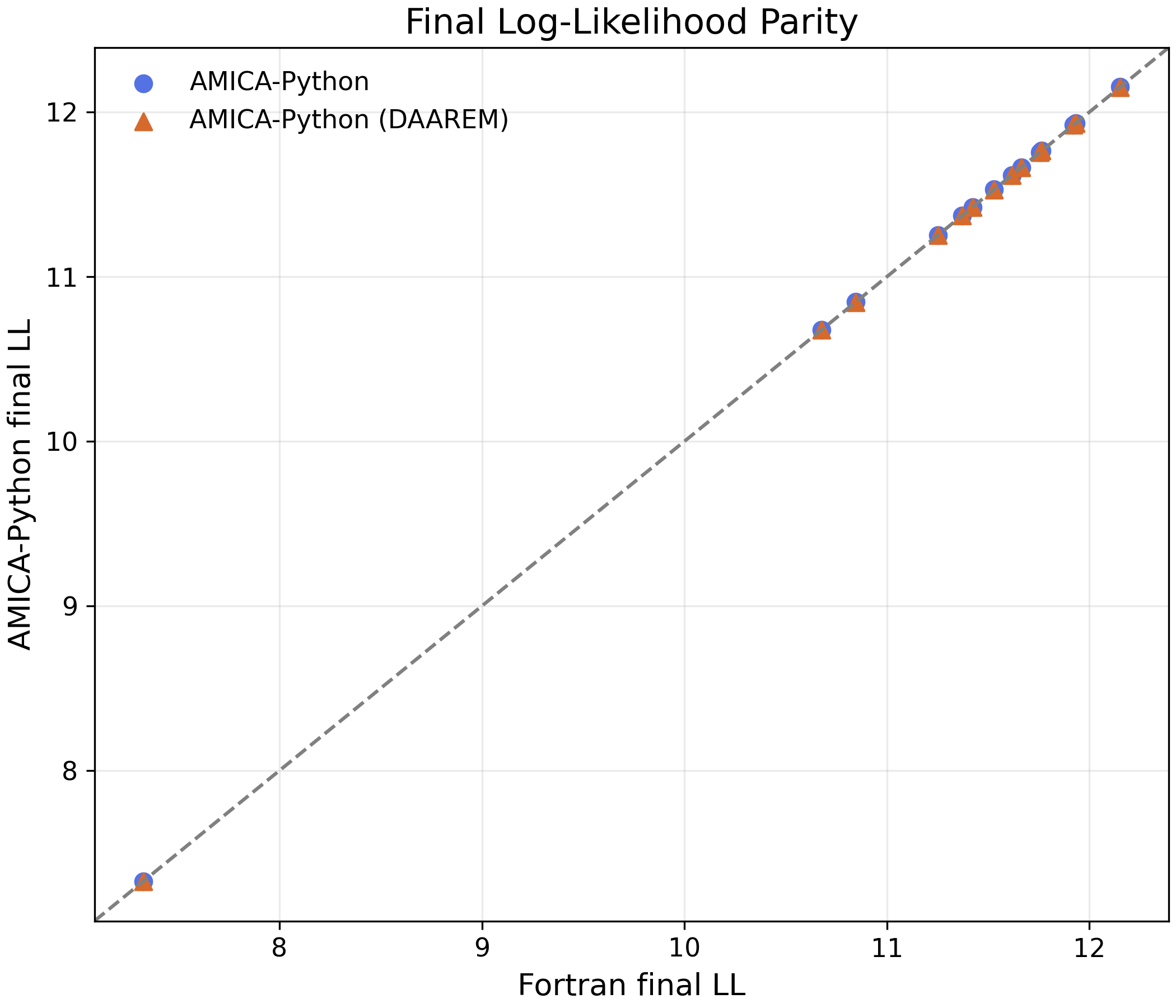}
\caption{Parity of the final normalized log-likelihood values obtained from the Fortran
reference implementation and AMICA-Python across the 14 benchmark recordings.}
\label{fig:benchmark-parity}
\end{figure}

The DAAREM-accelerated variant traded a small loss in final log-likelihood for
faster convergence: its median final normalized log-likelihood was 11.57267,
compared with 11.57285 for standard AMICA-Python EM. This corresponded to a
median relative absolute difference of $5.96\times10^{-6}$ from standard
AMICA-Python EM and $5.96\times10^{-6}$ from the Fortran reference. As shown in
\cref{fig:benchmark-parity}, standard AMICA-Python EM and the Fortran reference
were visually indistinguishable on most benchmark runs.

\subsection*{Replication of Component Mutual Information Reduction and Dipolarity analyses from Delorme (2004)}
\begin{figure}[t]
\centering
\includegraphics[width=\columnwidth]{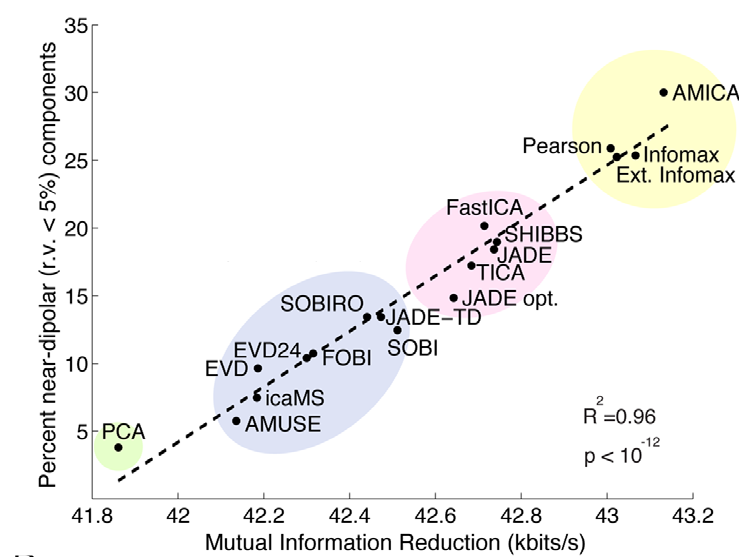}

\vspace{0.75em}

\includegraphics[width=\columnwidth]{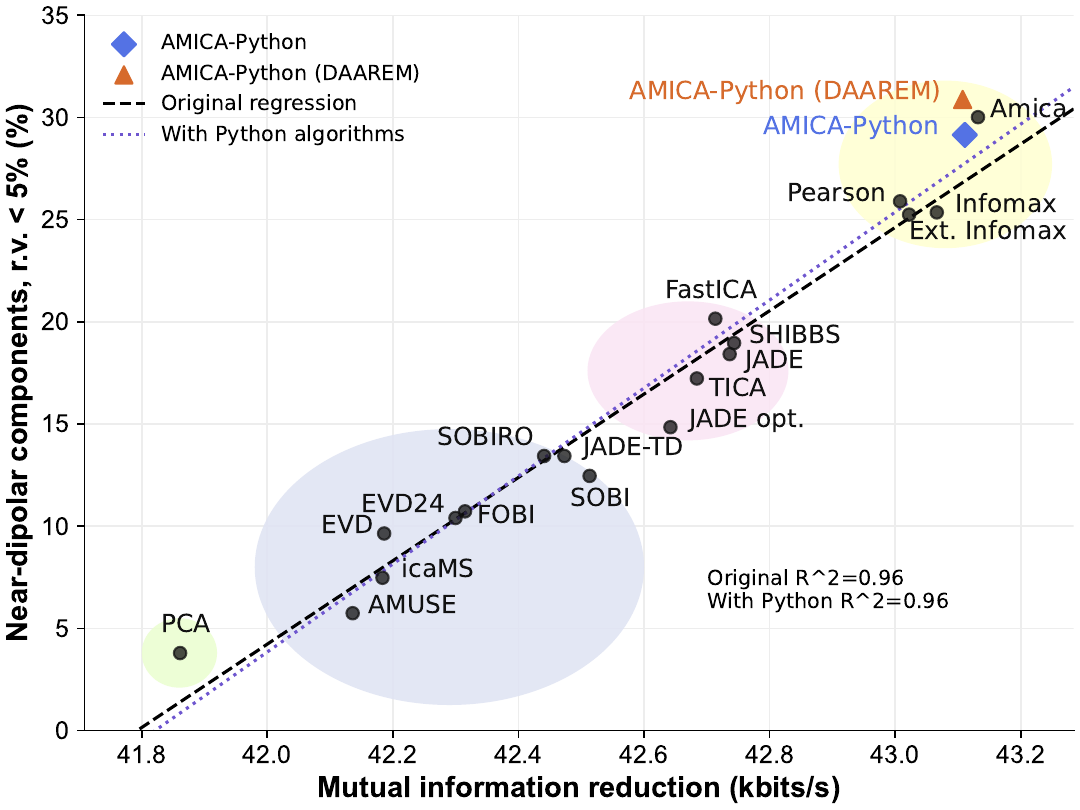}
\caption{Replication of Delorme et al.'s comparison between mutual information
reduction and the percentage of near-dipolar components. Top: the original published
panel from Delorme et al. Bottom: the recreated plot,
with AMICA-Python and its accelerated variant added as additional points. Mutual information reduction was
computed with the original MATLAB script and source dipolarity was estimated with the
original DIPFIT workflow, excluding recording \texttt{gv84}.}

\label{fig:delorme-figure4b-amica-python}
\end{figure}

The replicated Delorme analysis placed both AMICA-Python variants in the same
high-performing region as the original AMICA decomposition
(\cref{fig:delorme-figure4b-amica-python}). Across the 13 datasets used for this
analysis, mutual information reduction was nearly identical across implementations: $43.132 \pm 7.405$ kbits/s for Fortran AMICA, $43.111 \pm 7.400$ kbits/s for AMICA-Python, and $43.108 \pm 7.397$ kbits/s for the DAAREM-accelerated variant. Although the small MIR differences were statistically detectable in paired tests against Fortran (AMICA-Python: $p=0.028$; DAAREM: $p=0.015$), their magnitudes were less than $0.025$ kbits/s. The corresponding percentages of near-dipolar components were 29.14\% for AMICA-Python, 30.88\% with DAAREM
acceleration, and 30.01\% for AMICA Fortran, using the residual-variance threshold of
5\%. The percentage near-dipolar components in the Python implemenations did not differ significantly from Fortran ($p=0.491$ and $p=0.527$). Adding the Python decompositions to the original 18-algorithm scatter preserved the relationship between mutual information reduction and source
dipolarity: the original algorithms had $R^2=0.964$, and the augmented set
including AMICA-Python had $R^2=0.962$.s

  Across the 13 recordings used for this analysis, 

In terms of runtime, the Python implementation was competitive with the Fortran
reference. Across three matched benchmark runs, AMICA-Python averaged
82.3\% of the Fortran wall time and was faster than the reference on
all 14 datasets, even without acceleration. The DAAREM-accelerated variant averaged
65.9\% of the Fortran wall time and was faster than the reference on 13 of 14 datasets.
Runtime differences varied across datasets, and DAAREM was not uniformly faster than the
standard algorithm, but it reduced wall time for most datasets under the matched
single-core benchmark conditions (\cref{fig:benchmark-runtime}). Per-dataset
benchmark values are listed in the Appendix
(\cref{tab:benchmark-appendix,tab:benchmark-appendix-daarem,tab:benchmark-appendix-daarem-em}).

\begin{figure*}[t]
\centering
\includegraphics[width=\textwidth]{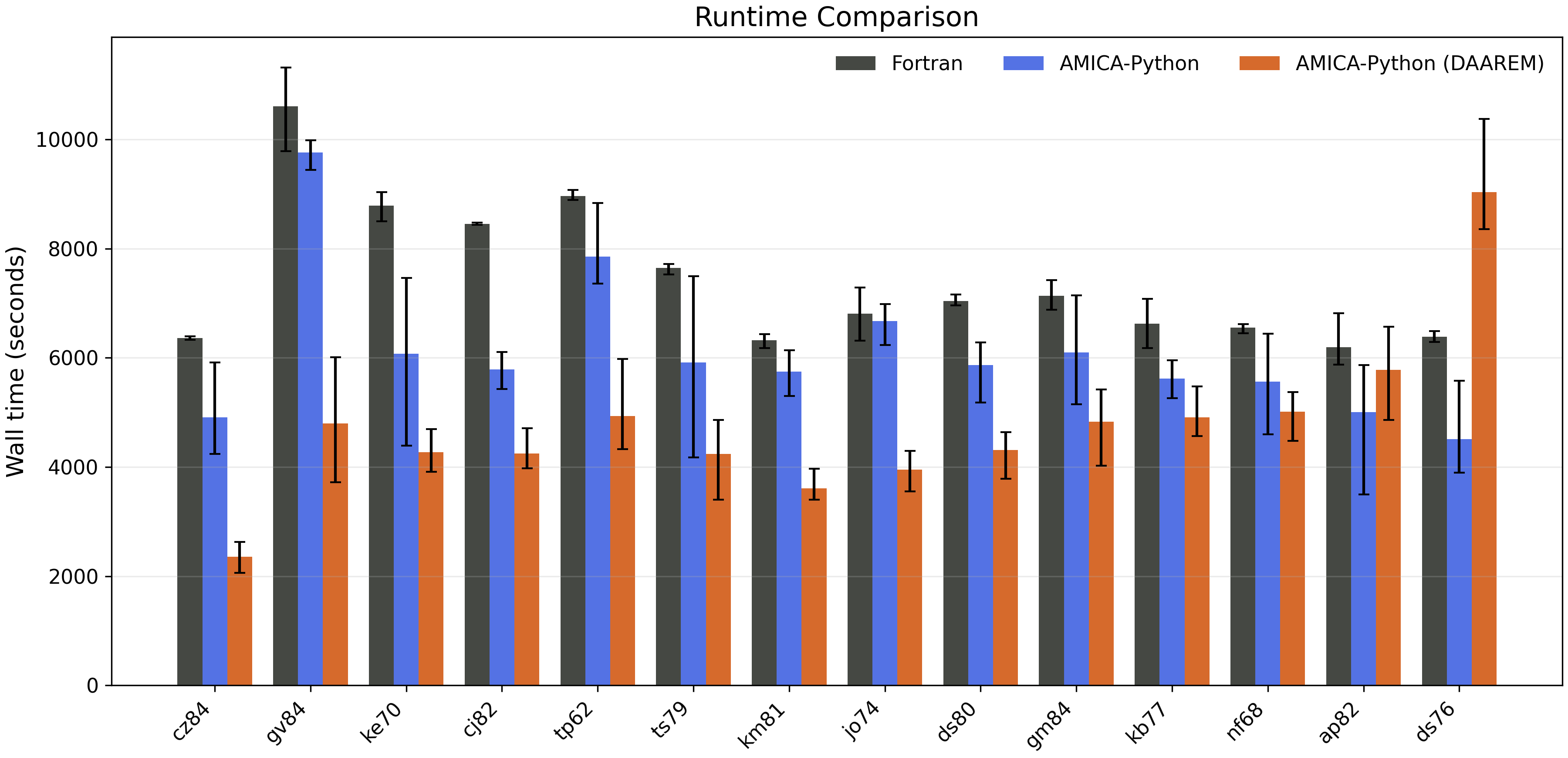}
\caption{Wall-clock runtime comparison between the Fortran reference
implementation and AMICA-Python across benchmark datasets. Bars show the
participant-level mean across three matched runs; error bars show the observed
minimum-to-maximum range across runs.}
\label{fig:benchmark-runtime}
\end{figure*}

Taken together, these results suggest that AMICA-Python reproduces the
reference implementation to high numerical precision on most datasets while
remaining competitive in runtime under matched benchmark settings. Further,
the addition of Anderson Acceleration has shown to be helpful for a variety of input
data, by dramatically reducing the time to fit the AMICA model.

\section*{Discussion}
AMICA has long been valued for its empirical performance, particularly in EEG
research, but in practice it has remained tied to a Fortran executable and its
EEGLAB wrapper. The goal of this work was to bring a robust and numerically
correct implementation of AMICA into the Python ecosystem, where much of modern
scientific computing and neurophysiological data analysis now takes place. To
that end, we presented AMICA-Python as an implementation that closely follows
the reference algorithm while offering a more familiar, Pythonic interface for
end users.

An important part of that goal was software usability. By exposing AMICA
through a scikit-learn-compliant estimator interface, AMICA-Python is easier to use
within existing analytical pipelines and more consistent with the conventions
already used in scientific Python tools, removing a significant technical hurdle that has blocked the integration of this algorithm in Python pipelines for many years. This lowers the barrier both for
applied users who want to incorporate AMICA into their analyses and for
developers who want to inspect and extend the method.

More broadly, this project aims to give AMICA a realistic opportunity to be
adopted beyond the narrower path of direct Fortran execution or EEGLAB-based
use. The benchmark results reported here suggest that this increased
accessibility comes with only negligible numerical differences when compared against
the reference implementation. We hope that making AMICA available as a native
Python tool will support wider use, easier comparison with other ICA
methods, and continued methodological development in the broader community.

Finally, we show that we can substantially improve upon the speed of the Fortran
implementation. While PyTorch certainly facilitates performant numerical processing, it
is not the sole driver of AMICA-Python's performance. While there is a tendency in modern
software engineering to deploy computationally expensive algorithms on GPU hardware or to compile them with numerical libraries such as PyTorch
or JAX, we found that improving the
efficiency of the AMICA algorithm (e.g. avoiding repeated or unnecessary operations) also greatly aided
our efforts to make AMICA-Python performant. Finally, like Pierre and colleagues
demonstrated with the extended-infomax algorithm \cite{ablin_faster_2017}, we found that
accelerating the optimization routine contributes much to speeding up the
AMICA algorithm.

\subsection*{Future directions} \label{sec:future}
Several extensions of AMICA-Python remain open for future work. In terms of software
implementation, broadening the support for alternative numerical
backends available in the Python ecosystem, e.g. NumPy, PyTorch, or JAX, would be a valuable extension. A natural next
step would be to rework the implementation around the Python Array API standard
\cite{meurer_python_2023} so that
users can choose among the supported numerical backends depending on their performance
and deployment needs. In principle, such an approach could make it easier to support a
wide base of users while preserving a common estimator interface.

Perhaps the most pressing barrier for future work, however, is optimization. In
practice, the Newton-related updates are among the most computationally
expensive parts of AMICA because they must repeatedly combine adaptive density
estimation with curvature-aware updates of the unmixing matrix. This cost is
especially important on the long, high-dimensional recordings for which AMICA
is often attractive. Related work on ICA optimization suggests that there may
be room for improvement on this front. Ablin and colleagues showed that preconditioned
quasi-Newton methods can substantially improve convergence speed on real ICA
problems by using Hessian approximations as preconditioners for L-BFGS rather
than relying only on first-order updates or on more expensive exact
second-order methods \cite{ablin_faster_2017}. Ille (2024)
similarly showed that alternative update parameterizations and curvature-aware
optimization strategies can markedly improve convergence behavior for
Infomax-style ICA on EEG data \cite{ille_orthogonal_2024}. For
AMICA-Python, this suggests several concrete directions: exploring better
preconditioning for the unmixing updates, considering alternative optimization strategies,
and testing whether hybrid quasi-Newton schemes can preserve AMICA's adaptive source
modeling while reducing the optimization bottleneck.

Finally, another direction concerns the source-density model itself. The current AMICA
formulation approximates each source density with a mixture of generalized
Gaussians, but this is not the only possible choice. As noted by Hsu and
colleagues, alternative density approximations could also be explored,
including other parametric families and more flexible nonparametric
approaches \cite{hsu_amica_2018}. While benchmarking alternative approaches to source
density estimation was out of scope for this project, we hope that making AMICA
available to the Python community will facilitate future research in this direction.
For example, one possible extension would be to make the
source-density estimation modular, allowing users to supply custom density
estimators within the AMICA framework.

\clearpage
\onecolumn
\appendix
\section*{Appendix: Benchmark Dataset Results}
The table below lists the per-recording benchmark values used to summarize
agreement and runtime in the main Results section.

{\scriptsize
\setlength{\tabcolsep}{3pt}
\begin{longtable}{@{}lrrrrrrrr@{}}
\caption{Per-dataset benchmark results for Fortran and Python (Py-EM). Values are participant-level means across three runs. The runtime ratio is Python divided by Fortran, so values below 1 indicate faster Python fits.}\label{tab:benchmark-appendix}\\
\toprule
Dataset & Fortran LL & Py-EM LL & $|\Delta \mathrm{LL}|$ & Fortran iter & Py-EM iter & Fortran s & Py-EM s & Py-EM/Ft \\
\midrule
\endfirsthead
\toprule
Dataset & Fortran LL & Py-EM LL & $|\Delta \mathrm{LL}|$ & Fortran iter & Py-EM iter & Fortran s & Py-EM s & Py-EM/Ft \\
\midrule
\endhead
\midrule
\multicolumn{9}{r}{Continued on next page}\\
\midrule
\endfoot
\bottomrule
\endlastfoot
ap82 & 11.251852 & 11.251852 & 1.33e-07 & 881.0 & 882.0 & 6190.0 & 5001.2 & 0.805 \\
cj82 & 11.616482 & 11.616483 & 1.50e-06 & 1206.0 & 1206.0 & 8450.1 & 5785.9 & 0.685 \\
cz84 & 11.422811 & 11.422812 & 8.07e-08 & 1006.0 & 1006.0 & 6364.0 & 4911.8 & 0.772 \\
ds76 & 11.370937 & 11.370938 & 1.10e-07 & 906.0 & 906.0 & 6381.1 & 4508.6 & 0.707 \\
ds80 & 11.755703 & 11.755703 & 7.72e-08 & 1006.0 & 1006.0 & 7036.9 & 5863.2 & 0.834 \\
gm84 & 7.328114 & 7.328114 & 4.98e-08 & 1006.0 & 1006.0 & 7132.0 & 6099.4 & 0.854 \\
gv84 & 10.846729 & 10.846767 & 3.80e-05 & 2000.0 & 2000.0 & 10602.9 & 9763.8 & 0.925 \\
jo74 & 11.764663 & 11.764679 & 1.64e-05 & 897.0 & 998.0 & 6804.8 & 6675.1 & 0.982 \\
kb77 & 10.677722 & 10.677714 & 7.48e-06 & 906.0 & 873.0 & 6627.8 & 5619.5 & 0.849 \\
ke70 & 11.933434 & 11.933422 & 1.20e-05 & 1306.0 & 1106.0 & 8787.3 & 6071.2 & 0.693 \\
km81 & 11.529211 & 11.529211 & 7.42e-08 & 906.0 & 906.0 & 6317.9 & 5748.1 & 0.909 \\
nf68 & 11.663512 & 11.663512 & 8.29e-08 & 948.0 & 946.0 & 6549.8 & 5561.9 & 0.849 \\
tp62 & 11.920194 & 11.920195 & 1.28e-06 & 1299.0 & 1300.0 & 8965.8 & 7855.6 & 0.877 \\
ts79 & 12.151492 & 12.151492 & 3.39e-08 & 1106.0 & 1106.0 & 7643.2 & 5916.1 & 0.776 \\
\end{longtable}
}

{\scriptsize
\setlength{\tabcolsep}{3pt}
\begin{longtable}{@{}lrrrrrrrr@{}}
\caption{Per-dataset benchmark results for Fortran and Python with DAAREM accelleration (Py-DAAREM). Values are participant-level means across three matched runs. The runtime ratio is Python divided by Fortran, so values below 1 indicate faster Python fits.}\label{tab:benchmark-appendix-daarem}\\
\toprule
Dataset & Fortran LL & Py-DAAREM LL & $|\Delta \mathrm{LL}|$ & Fortran iter & Py-DAAREM iter & Fortran s & Py-DAAREM s & Py-DAAREM/Ft \\
\midrule
\endfirsthead
\toprule
Dataset & Fortran LL & Py-DAAREM LL & $|\Delta \mathrm{LL}|$ & Fortran iter & Py-DAAREM iter & Fortran s & Py-DAAREM s & Py-DAAREM/Ft \\
\midrule
\endhead
\midrule
\multicolumn{9}{r}{Continued on next page}\\
\midrule
\endfoot
\bottomrule
\endlastfoot
ap82 & 11.251852 & 11.251828 & 2.42e-05 & 881.0 & 891.0 & 6190.0 & 5779.4 & 0.932 \\
cj82 & 11.616482 & 11.616501 & 1.89e-05 & 1206.0 & 906.0 & 8450.1 & 4243.2 & 0.502 \\
cz84 & 11.422811 & 11.422540 & 2.72e-04 & 1006.0 & 406.0 & 6364.0 & 2352.3 & 0.370 \\
ds76 & 11.370937 & 11.370865 & 7.26e-05 & 906.0 & 1706.0 & 6381.1 & 9033.6 & 1.416 \\
ds80 & 11.755703 & 11.755638 & 6.50e-05 & 1006.0 & 651.0 & 7036.9 & 4311.6 & 0.613 \\
gm84 & 7.328114 & 7.327969 & 1.45e-04 & 1006.0 & 706.0 & 7132.0 & 4829.2 & 0.678 \\
gv84 & 10.846729 & 10.844748 & 1.98e-03 & 2000.0 & 1137.0 & 10602.9 & 4801.0 & 0.458 \\
jo74 & 11.764663 & 11.764476 & 1.87e-04 & 897.0 & 506.0 & 6804.8 & 3949.6 & 0.581 \\
kb77 & 10.677722 & 10.677702 & 1.98e-05 & 906.0 & 648.0 & 6627.8 & 4908.9 & 0.742 \\
ke70 & 11.933434 & 11.933440 & 5.95e-06 & 1306.0 & 918.0 & 8787.3 & 4268.8 & 0.486 \\
km81 & 11.529211 & 11.528832 & 3.80e-04 & 906.0 & 506.0 & 6317.9 & 3610.3 & 0.571 \\
nf68 & 11.663512 & 11.663523 & 1.11e-05 & 948.0 & 705.0 & 6549.8 & 5009.3 & 0.765 \\
tp62 & 11.920194 & 11.920212 & 1.85e-05 & 1299.0 & 948.0 & 8965.8 & 4932.0 & 0.550 \\
ts79 & 12.151492 & 12.150648 & 8.44e-04 & 1106.0 & 806.0 & 7643.2 & 4242.2 & 0.556 \\
\end{longtable}
}

{\scriptsize
\setlength{\tabcolsep}{3pt}
\begin{longtable}{@{}lrrrrrrrr@{}}
\caption{Per-dataset comparison between The Python implementation (Py-EM) and its accelerated variant (Py-DAAREM). Values are participant-level means across three matched runs. The runtime ratio is Py-DAAREM divided by Py-EM, so values below 1 indicate faster DAAREM fits.}\label{tab:benchmark-appendix-daarem-em}\\
\toprule
Dataset & Py-EM LL & Py-DAAREM LL & DAAREM $-$ EM LL & Py-EM iter & Py-DAAREM iter & Py-EM s & Py-DAAREM s & DAAREM/EM \\
\midrule
\endfirsthead
\toprule
Dataset & Py-EM LL & Py-DAAREM LL & DAAREM $-$ EM LL & Py-EM iter & Py-DAAREM iter & Py-EM s & Py-DAAREM s & DAAREM/EM \\
\midrule
\endhead
\midrule
\multicolumn{9}{r}{Continued on next page}\\
\midrule
\endfoot
\bottomrule
\endlastfoot
ap82 & 11.251852 & 11.251828 & -2.43e-05 & 882.0 & 891.0 & 5001.2 & 5779.4 & 1.186 \\
cj82 & 11.616483 & 11.616501 & 1.74e-05 & 1206.0 & 906.0 & 5785.9 & 4243.2 & 0.733 \\
cz84 & 11.422812 & 11.422540 & -2.72e-04 & 1006.0 & 406.0 & 4911.8 & 2352.3 & 0.482 \\
ds76 & 11.370938 & 11.370865 & -7.27e-05 & 906.0 & 1706.0 & 4508.6 & 9033.6 & 2.024 \\
ds80 & 11.755703 & 11.755638 & -6.50e-05 & 1006.0 & 651.0 & 5863.2 & 4311.6 & 0.735 \\
gm84 & 7.328114 & 7.327969 & -1.45e-04 & 1006.0 & 706.0 & 6099.4 & 4829.2 & 0.797 \\
gv84 & 10.846767 & 10.844748 & -2.02e-03 & 2000.0 & 1137.0 & 9763.8 & 4801.0 & 0.492 \\
jo74 & 11.764679 & 11.764476 & -2.03e-04 & 998.0 & 506.0 & 6675.1 & 3949.6 & 0.591 \\
kb77 & 10.677714 & 10.677702 & -1.23e-05 & 873.0 & 648.0 & 5619.5 & 4908.9 & 0.873 \\
ke70 & 11.933422 & 11.933440 & 1.80e-05 & 1106.0 & 918.0 & 6071.2 & 4268.8 & 0.733 \\
km81 & 11.529211 & 11.528832 & -3.79e-04 & 906.0 & 506.0 & 5748.1 & 3610.3 & 0.628 \\
nf68 & 11.663512 & 11.663523 & 1.10e-05 & 946.0 & 705.0 & 5561.9 & 5009.3 & 0.910 \\
tp62 & 11.920195 & 11.920212 & 1.73e-05 & 1300.0 & 948.0 & 7855.6 & 4932.0 & 0.637 \\
ts79 & 12.151492 & 12.150648 & -8.44e-04 & 1106.0 & 806.0 & 5916.1 & 4242.2 & 0.733 \\
\end{longtable}
}

\begin{acknowledgements}
placeholder for funding, institutional support, and project acknowledgements.
\end{acknowledgements}

\begin{contributions}
Scott Huberty carried out the re-implementation of the AMICA program in Python, conducted
the benchmark analysis, and contributed to the manuscript preparation. Christian O'Reilly provided guidance on project diretion, reviewed and edited the manuscript, providing comments and advice.
\end{contributions}

\begin{interests}
The authors declare no competing interests.
\end{interests}

\section*{Bibliography}
\bibliography{references}

\end{document}